\pdfoutput=1

\documentclass[11pt]{article}

\usepackage[preprint]{acl}

\usepackage{times}
\usepackage{latexsym}

\usepackage{inconsolata}

\usepackage[utf8]{inputenc} 
\usepackage[T1]{fontenc}    
\usepackage{hyperref}       
\usepackage{url}            
\usepackage{booktabs}       
\usepackage{amsfonts}       
\usepackage{nicefrac}       
\usepackage{microtype}      
\usepackage{xcolor}         

\definecolor{Blue9}{rgb}{0.098,0.3,0.9}

\usepackage{amsmath,amsfonts,bm}

\def\eqref#1{Eq.~(\ref{#1})}

\def\1{\bm{1}}

\def\vl{{\bm{l}}}

\DeclareMathAlphabet{\mathsfit}{\encodingdefault}{\sfdefault}{m}{sl}
\SetMathAlphabet{\mathsfit}{bold}{\encodingdefault}{\sfdefault}{bx}{n}

\def\gA{{\mathcal{A}}}

\def\gD{{\mathcal{D}}}

\def\gM{{\mathcal{M}}}
\def\gN{{\mathcal{N}}}

\def\gP{{\mathcal{P}}}

\def\gR{{\mathcal{R}}}
\def\gS{{\mathcal{S}}}
\def\gT{{\mathcal{T}}}
\def\gU{{\mathcal{U}}}

\newcommand{\E}{\mathbb{E}}

\newcommand{\ourmethod}{VLP\xspace}
\newcommand{\ourdata}{MTVLP\xspace}
\newcommand{\MW}{Meta-World\xspace}
\newcommand{\MS}{ManiSkill2\xspace}

\newcommand{\prefmodel}{f_\psi}

\newcommand{\vinput}{v}
\newcommand{\vembed}{z}

\newcommand{\linput}{l}
\newcommand{\lembed}{u}

\newcommand{\CE}{\operatorname{CE}}

\newcommand{\PbRL}{Preference-based RL\xspace}
\newcommand{\pbRL}{preference-based RL\xspace}
\newcommand{\RL}{Reinforcement Learning\xspace}

\newcommand{\lc}{language-conditioned\xspace}

\renewcommand{\vl}{vision-language\xspace}

\newcommand{\mean}[1]{#1}
\newcommand{\bmean}[1]{\textbf{#1}}
\newcommand{\std}[1]{\scalebox{.70}{$\pm$ #1}}

\newcommand{\mycolor}{cyan!10}
\newcommand{\highlight}{\cellcolor{\mycolor}}

\usepackage{amsmath}
\usepackage{amssymb}
\usepackage{mathtools}
\usepackage{amsthm}
\usepackage{mathtools} 

\theoremstyle{plain}

\theoremstyle{definition}

\theoremstyle{remark}

\usepackage{algorithm}
\usepackage{algorithmicx}
\usepackage{algpseudocode}

\usepackage{enumitem} 
\usepackage{subfig} 
\usepackage{multirow}

\usepackage{xspace}
\usepackage{wrapfig}  
\usepackage{makecell} 

\usepackage{graphicx} 
\usepackage{colortbl}
\usepackage{tcolorbox}

\graphicspath{{figs/}}

\renewcommand{\mid}[0]{|}

\newcommand{\RC}{RoboCLIP\xspace}

\title{VLP: Vision-Language Preference Learning for Embodied Manipulation}

\author{
Runze Liu\textsuperscript{$1$},
Chenjia Bai\textsuperscript{$2$},
Jiafei Lyu\textsuperscript{$1$},
Shengjie Sun\textsuperscript{$1$},
Yali Du\textsuperscript{$3$},
Xiu Li\textsuperscript{$1$} \\
\textsuperscript{$1$}Tsinghua Shenzhen International Graduate School, Tsinghua University \\
\textsuperscript{$2$}Institute of Artificial Intelligence (TeleAI), China Telecom,
\textsuperscript{$3$}King’s College London
}

\begin{document}
\maketitle
\begin{abstract}
Reward engineering is one of the key challenges in Reinforcement Learning (RL). Preference-based RL effectively addresses this issue by learning from human feedback. However, it is both time-consuming and expensive to collect human preference labels. In this paper, we propose a novel \textbf{V}ision-\textbf{L}anguage \textbf{P}reference learning framework, named \textbf{VLP}, which learns a vision-language preference model to provide preference feedback for embodied manipulation tasks. To achieve this, we define three types of language-conditioned preferences and construct a vision-language preference dataset, which contains versatile implicit preference orders without human annotations. The preference model learns to extract language-related features, and then serves as a preference annotator in various downstream tasks. The policy can be learned according to the annotated preferences via reward learning or direct policy optimization. Extensive empirical results on simulated embodied manipulation tasks demonstrate that our method provides accurate preferences and generalizes to unseen tasks and unseen language instructions, outperforming the baselines by a large margin.
\end{abstract}

\section{Introduction}

Reinforcement Learning (RL) has made great achievements recent years, including board games~\citep{AlphaGo-Zero, AlphaZero}, autonomous driving~\citep{kiran2021deep, Smarts}, and robotic manipulation~\citep{kober2013reinforcement, andrychowicz2020learning, Bi-DexHands, sun2024enhancing, bai2025retrdex}. However, one of the key challenges to apply RL algorithms is reward engineering. First, designing an accurate reward function requires large amount of expert knowledge. Second, the agent might hack the designed reward function~\citep{IRD}, obtaining high returns without completing the task. Also, it is difficult to obtain reward functions for subjective human objectives.

To address the above issues, a variety of works leverage expert demonstrations for imitation learning (IL)~\citep{GAIL, GAIfO}. Nevertheless, expert demonstrations are often expensive and the performance of IL is limited by the quality of the demonstrations. Another line of work leverages Vision-Language Models (VLMs) to provide multi-modal rewards for downstream policy learning~\citep{R3M, LIV, VLM-RM}. However, the reward labels produced in these works are often of high variance and noisy~\citep{LIV}. Preference-based RL is more promising way that learns from human preferences over trajectory pairs~\citep{christiano2017deep, PEBBLE, COPO}. On the one hand, we can learn a reward model from preferences and then optimize the policy according to the reward model~\citep{christiano2017deep, PT}. On the other hand, the policy can be directly optimized according to the preferences~\citep{IPL, CPL}.

However, preference-based RL requires either querying a large number of expert preference labels online~\citep{PEBBLE, SURF} or a labeled offline preference dataset~\citep{PT, CPL}, which is quite time-consuming and expensive. As the reasoning abilities of Large Language Models~(LLMs) improve significantly~\citep{o1, liu2025can}, previous methods propose to use LLMs/VLMs to provide preference labels~\citep{PrefCLM, RL-VLM-F}, but the generated labels are not guaranteed to be accurate and it is assumed to have access to the environment information.

\begin{figure*}[t]
\centering
\vspace{-1em}
\includegraphics[width=1.0\linewidth]{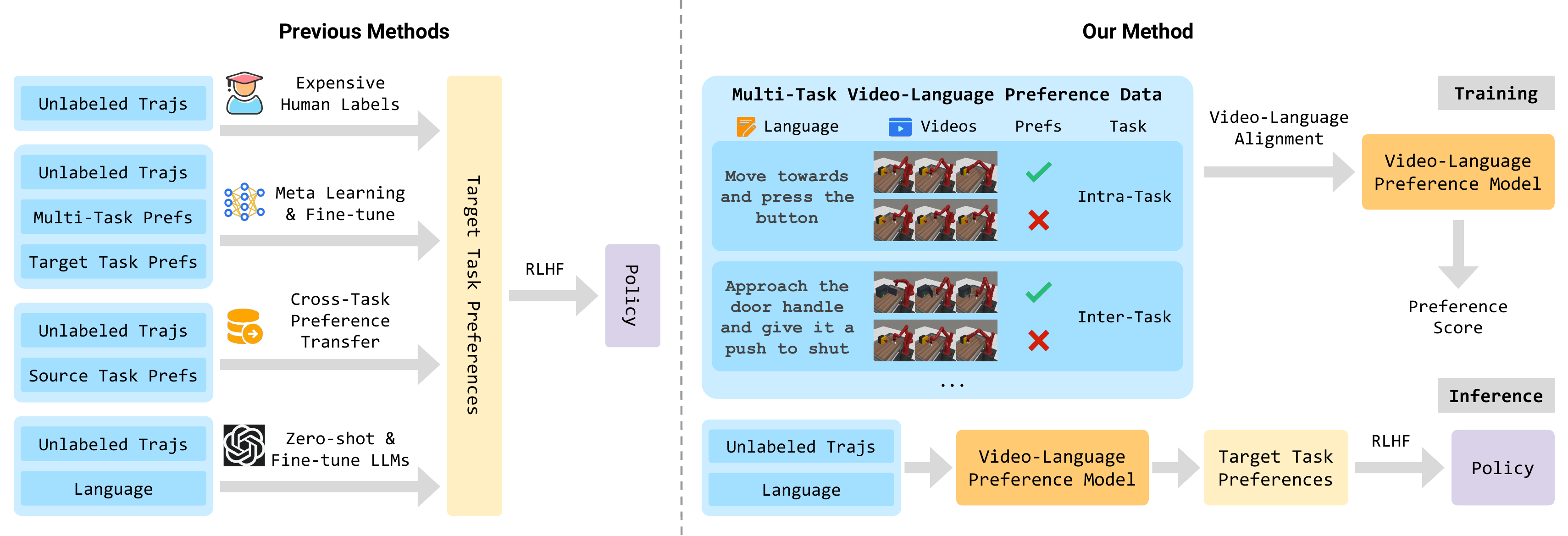}
\caption{Comparison of \ourmethod (right) with previous methods (left) of providing preference labels.
}
\vspace{-1em}
\label{fig:framework}
\end{figure*}

In this paper, we propose a novel \textbf{V}ision-\textbf{L}anguage \textbf{P}reference alignment framework, named \textbf{VLP}, to provide preference feedback for video pairs given language instructions. Specifically, we collect a video dataset from various policies under augmented language instructions, which contains implicit preference relations based on the trajectory optimality and the \vl correspondence. Then, we define \lc preferences and propose a novel vision-language alignment architecture to learn a trajectory-wise preference model for preference labeling, which consists of a video encoder, a language encoder, and a cross-modal encoder to facilitate \vl alignment. The preference model is optimized by intra-task and inter-task preferences that are implicitly contained in the dataset. In inference, \ourmethod provides preference labels for target tasks and can even generalize to unseen tasks and unseen language instructions. We provide an analysis to show the learned preference model resembles the negative regret of the segment under mild conditions. The preference labels given by \ourmethod are employed for various downstream preference optimization algorithms to facilitate policy learning.

In summary, our contributions are as follows:
(\romannumeral1) We propose a novel vision-language preference alignment framework, which learns a vision-language preference model to provide preference feedback for embodied manipulation tasks.
(\romannumeral2) We propose \lc preferences and construct a vision-language preference dataset, which contains videos with language instructions and implicit \lc relations.
(\romannumeral3) Extensive empirical results on simulated embodied manipulation tasks demonstrate that our method provides accurate preferences and generalizes to unseen tasks and unseen language instructions, outperforming the baselines by a large margin.

\section{Background}\label{sec:background}

\paragraph{Problem Setting.}
We formulate the RL problem as a Markov Decision Process (MDP)~\citep{sutton2018reinforcement} represented as a tuple $\gM = (\gS, \gA, \gP, \gR, \gamma, p_0)$, where $\gS$ is the state space, $\gA$ is the action space, $\gP: \gS \times \gA \to \gS$ is the transition function, $\gR: \gS \times \gA \to \mathbb{R}$ is the reward function, $\gamma \in [0, 1)$ is the discount factor, and $p_0: \gS \to [0, 1]$ is the initial state distribution.
At timestep $t$, the agent observes a state $s_t$ and selects an action $a_t$ based on a policy $\pi(a_t \mid s_t)$. Then, the agent receives a reward $r_t$ from the environment, and the agent transits to $s_{t+1}$ according to the transition function. The agent's goal is to find a policy that maximizes the expected cumulative reward $\mathbb{E} \left[\sum_{t=0}^\infty \gamma^t r_t \right]$. In multi-task setting, for a task $\gT \sim p(\gT)$, a task-specific MDP is represented as $\gM^{\gT} = (\gS^{\gT}, \gA, \gP^{\gT}, \gR^{\gT}, \gamma, p_0^{\gT})$.

\begin{figure*}[t]
\centering
\vspace{-1em}
\includegraphics[width=0.85\linewidth]{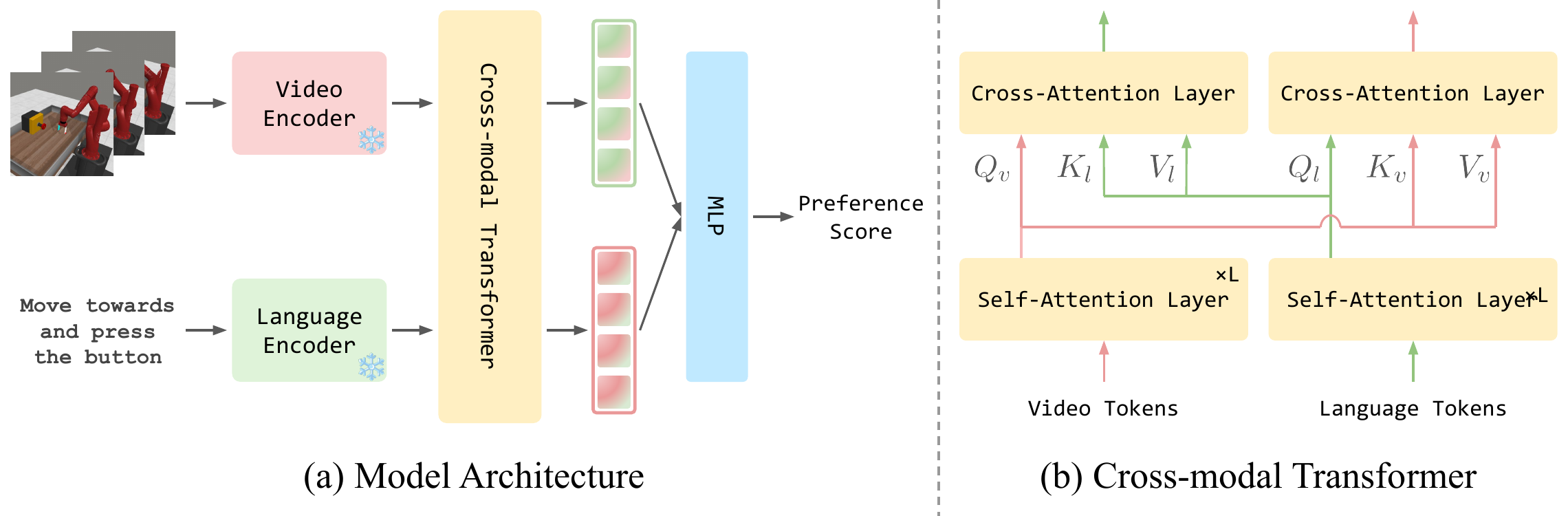}
\vspace{-0.5em}
\caption{(a) Trajectory videos and language instruction are fed into the preference model to obtain a trajectory-wise preference score. 
(b) The cross-modal transformer obtains language-related video features and video-related language features by cross-attention mechanism.
}
\vspace{-1em}
\label{fig:model}
\end{figure*}

\paragraph{Preference-based RL.}
\PbRL differs from RL in that it is assumed to have no access to the ground-truth rewards~\citep{christiano2017deep, PEBBLE}. In \pbRL, human teachers provide preference labels over trajectory pairs, and a reward model is learned from these preferences. Formally, a trajectory segment $\sigma$ of length $H$ is represented as $\{s_1, a_1, \ldots, s_H, a_H\}$ and a segment pair is $(\sigma^1, \sigma^2)$. The preference label $y \in \{0, 1, 0.5\}$ denotes which segment is preferred, where $0$ indicates $\sigma^1$ is preferred (i.e., $\sigma^1 \succ \sigma^2$), $1$ indicates $\sigma^2$ is preferred (i.e., $\sigma^2 \succ \sigma^1$), and $0.5$ represents two segments are equally preferred. Previous \pbRL approaches construct a preference predictor with the reward model $\widehat{r}_\psi$ via Bradley-Terry model~\citep{BTModel}:
\begin{equation}
    P_\psi[\sigma^1 \succ \sigma^2] = \frac{\exp \big(\sum_{t=1}^H \widehat{r}_\psi(s_t^1, a_t^1) \big)}{\sum_{k=1}^{2}\exp \big(\sum_{t=1}^H \widehat{r}_\psi(s_t^k, a_t^k) \big)},
\end{equation}
where $P_\psi[\sigma^1 \succ \sigma^2]$ denotes the probability that $\sigma^1$ is preferred over $\sigma^2$ predicted by current reward model $\widehat{r}_\psi$.
Assume we have a dataset with preference labels $\gD = \{(\sigma^1, \sigma^2, y)\}$, the reward learning process can be formulated as a classification problem using cross-entropy loss~\citep{christiano2017deep}:
\begin{equation}
\begin{aligned}
    \mathcal{L}_{\text{ce}} = - \underset{{(\sigma^1, \sigma^2, y) \sim \mathcal{D}}}{\E}
    & \Big[ (1-y) \log P_\psi[\sigma^1 \succ \sigma^2] \\
    &+ y \log P_\psi[\sigma^2 \succ \sigma^1] \Big].
\label{eq:sup_loss}
\end{aligned}
\end{equation}
By optimizing \eqref{eq:sup_loss}, the reward model is aligned with human preferences, providing reward signals for policy learning.

\section{Method}

In this section, we first present the overall framework of \ourmethod, including model architecture and the \vl preference dataset. Then, we introduce \lc preferences and the detailed algorithm for \vl preference learning, which learns a trajectory-wise preference model via \vl preference alignment.

\subsection{Model and Dataset}
\label{sec:framwork}

The goal of \ourmethod is to learn a generalized preference model capable of providing preferences for novel embodied tasks. To achieve this, the preference model receives videos and language as inputs, where videos serve as universal representations of agent trajectories and language act as universal and flexible instructions.
To obtain high-quality representations of these two modalities, we utilize CLIP~\citep{CLIP}, which is pre-trained on extensive image-text data, as our video and language encoders. The extracted video and language features are fed into to a cross-modal transformer for cross-modal attention interaction to capture video features associated with the language and language features related to the video. These features are subsequently utilized for predicting preference scores in \vl preference learning. The overall framework is illustrated in Figure~\ref{fig:model}.

\paragraph{Model Architecture.}

A video $\vinput$ is represented as a sequence of video frames, i.e., $\vinput = \{\vinput_1, \vinput_2, \ldots, \vinput_{|\vinput|}\}$, where $\vinput_i \in \mathbb{R}^{H \times W \times 3}$, $H$ and $W$ are the height and width of each video frame, and $|\vinput|$ denotes the number of video frames. The video encoder is employed to obtain the video tokens $\vembed = \{\vembed_1, \vembed_2, \ldots, \vembed_{|\vinput|}\}$, where $\vembed_i \in \mathbb{R}^{M \times D_v}$, $M = H / p \times W / p$ is the number of visual tokens, $p$ is the patch size of CLIP ViT, and $D_v$ is the dimension of the visual tokens. Given language input $\linput$, the language tokens $\lembed \in \mathbb{R}^{N \times D_l}$ are obtained via the language encoder, where $N$ is the number of language tokens, and $D_l$ is the dimension of the language tokens.

With video tokens $\vembed$ and language tokens $\lembed$, a cross-modal encoder is employed to facilitate multi-modal feature learning, making tokens of different modalities fully fuse with each other.
Video tokens and language tokens are separately inputted into the self-attention layers.
Then, utilizing the output video tokens as queries and the output language tokens as keys and values, the cross-attention layer, as shown in Figure~\ref{fig:model}(b), generates language features that are closely related to the input video. Similarly, the cross-attention layer produces language-related video features.
The multi-modal tokens are averaged along the first dimension and then concatenated as $w \in \mathbb{R}^{D_w}$, where $D_w = D_v+D_l$. These new tokens are fed into the final Multi-layer Perceptron (MLP) for \vl preference prediction, outputting a trajectory-level preference score.

\paragraph{Vision-Language Preference Dataset.}

While there are open-sourced embodied datasets with language instructions~\citep{EmbodiedGPT}, there lacks a multi-modal preference dataset for generalized preference learning.
To this end, we construct \ourdata, a multi-task \vl preference dataset built upon \MW~\citep{Meta-World}.
To that end, we consider the following aspects:
(\romannumeral1)~trajectories of various optimality levels should be collected to define clear preference relations within each task;
(\romannumeral2)~each trajectory pair should be accompanied with a corresponding language instruction for learning \lc preferences.

It is easy to describe the optimality of expert trajectories and random trajectories because it is easy to understand the agent's behavior in these trajectories.
However, it is challenging to define a medium-level policy without explicit rewards. Fortunately, we find most robot tasks can be divided into multiple stages, where each stage completes a part of the overall task. Thus, we define a medium-level policy as successfully completing half of the stages of the task.
For example, we divided the task of \emph{opening the drawer} into two subtasks: (\romannumeral1) moving and grasping the drawer handle and (\romannumeral2) pulling the drawer handle. A medium-level policy only completes the first subtask.

We leverage a scripted policy for each task to roll out trajectories of three optimality levels: expert, medium, and random. For expert-level trajectories, 
we employ the scripted policy with Gaussian noise to interact.
The medium-level trajectories are also collected with the scripted policy but are terminated when the half of subtasks are completed.
As for random-level trajectories, actions are randomly sampled from a uniform distribution during rollout.
For the corresponding language, we obtain diverse language instructions to improve the generalization abilities of our model by aligning one video with multiple similar language instructions. Following~\citet{LAMP}, we query GPT-4V~\citep{GPT-4V} to generate language instructions with various verb structure examples and synonym nouns of each task.
Details of collecting trajectories and language instructions for each task are shown in Appendix~\ref{app:data_details}.

\subsection{Vision-Language Preference Alignment}
\label{sec:vlalign}

\paragraph{Language-conditioned Preferences.}
Previous RLHF methods define trajectory preferences according to a single task goal. However, this uni-modal approach struggles to generalize to new tasks due to its rigid preference definition. In contrast, by integrating language as a condition, we can establish more flexible preference definitions. Consider two videos, $\vinput_1^1$ and $\vinput_2^1$, along with a language instruction $\linput^1$ from task $\gT^1$, and another video $\vinput^2$ paired with a language instruction $\linput^2$ from task $\gT^2$. We categorize three forms of language-conditioned preferences: Intra-Task Preference (ITP), Inter-Language Preference (ILP), and Inter-Video Preference (IVP), as shown in Table~\ref{tab:preference}.

\begin{table}[!htbp]
\centering
\caption{Three types of language-conditioned preferences.}
\resizebox{1.0\linewidth}{!}{
\begin{tabular}{llll}
\toprule
{\bf Type} & {\bf Videos} & {\bf Language} & {\bf Criterion} \\
\midrule
ITP & $\vinput_1^1, \vinput_2^1 \sim \gT^1$ & $\linput^1 \sim \gT^1$ & optimality \\
ILP & $\vinput_1^1, \vinput_2^1 \sim \gT^1$ & $\linput^2 \sim \gT^2$ & equally preferred \\
IVP & $\vinput_1^1 \sim \gT^1$, $\vinput_1^2 \sim \gT^2$ & $\linput^1 \sim \gT^1$ & $\vinput_1^1 \succ \vinput_2^1 \mid \linput^1$ \\
\bottomrule
\end{tabular}
}
\label{tab:preference}
\end{table}

ITP corresponds to the conventional case of preference relation within the same task~\citep{christiano2017deep}, where the videos and language instructions are from the same task, and the preference relies on the optimality of videos w.r.t. the task objective.
ILP considers a scenario where the language instruction differs from the task of the videos. Thus, both videos are equally preferred under this language condition.
IVP deals with preferences of two videos from different tasks, with the language instruction from either task. It is straightforward to define the preference that the \vl come from the same task is preferred to the other pair.

This framework allows for the establishment of universal and adaptable preference relations, wherein videos from the same task can yield varying preference labels depending on the language condition.
Notably, even random trajectories paired with language instructions from a specific task is preferred to expert trajectories from other tasks.

\paragraph{Vision-Language Preference Learning.}

With \lc preferences defined above, we further introduce our \vl preference learning algorithm. We aim to develop a \vl preference model that predicts the preferred video under specific language conditions. However, directly inputting two videos and a language instruction into the model would affect computational efficiency. So, we consider the conventional way to learn from preference labels~\citep{christiano2017deep}, i.e., first constructing preference predictors via Bradley-Terry model~\citep{BTModel}. Previous work has revealed the advantages of learning a preference model over a reward model~\citep{FTB}. Based on these insights, our proposed preference model $\prefmodel(\vinput \mid \linput)$ takes a video and a language instruction as inputs and outputs a scalar preference score. Then the preference label can be obtained by comparing preference scores of two videos with a given language instruction, i.e., $\vinput_1 \succ \vinput_2 | \linput$ if $\prefmodel(\vinput_1 | \linput) > \prefmodel(\vinput_2 | \linput)$.

Given videos $\vinput_1$ representing $\sigma_1$ and $\vinput_2$ representing $\sigma_2$, the \lc preference distribution $P_\psi[\vinput_1 \succ \vinput_2 \mid \linput]$ is the probability that $\sigma_1$ is preferred over $\sigma_2$ under the condition $\linput$:
\begin{equation}
    P_\psi[\vinput_1 \succ \vinput_2 \mid \linput] = \frac{\exp \big( \prefmodel(\vinput_1 \mid \linput) \big)}{\sum_{k=1}^{2}\exp \big( \prefmodel(\vinput_k \mid \linput) \big)}.
\label{eq:lc_BT}
\end{equation}

Given tasks $\gT^1$ and $\gT^2$, we consider the following objectives aligned with \lc preference relations:
\textbf{(a)} Learning Intra-Task Preference: Within the same task, the video that better follows $l$ should be preferred, analogous to previous RLHF objective~\citep{christiano2017deep};
\textbf{(b)} Learning Inter-Language Preference: Under the language condition of task $\gT^2$, videos from task $\gT^1$ are equally preferred;
\textbf{(c)} Learning Inter-Video Preference: Under the language condition of task $\gT^1$, the video from $\gT^1$ is preferred over the video from $\gT^2$.

During \vl preference learning, a task $\gT$ is sampled from all training tasks, followed by sampling a minibatch $\{\vinput_1^b, \vinput_2^b, \vinput^{\neq b}, \linput^b, \linput^{\neq b}, y^{\text{ITP}}, y^{\text{ILP}}, y^{\text{IVP}} \}^{1:B}$.
Here, the superscript $^b$ indicates data sampled from task $\gT$ in the minibatch, while $^{\neq b}$ denotes data from other tasks.
$y^{\text{ITP}}, y^{\text{ILP}}, y^{\text{IVP}}$ are the ground-truth labels of ITP, ILP, and IVP, respectively.
The total loss of \vl preference learning is as follows:
\begin{equation}
\begin{aligned}
    \mathcal{L}_{\text{ce}} = - \sum_{b\in B}
    \Big[
    & \underbrace{\CE\left( P_\psi[\vinput_1^b \succ \vinput_2^b \mid \linput^b], y^{\text{ITP}} \right)}_{\text{(a)}} \\
    &+ \lambda_1 \underbrace{\CE\left( P_\psi[\vinput_1^b \succ \vinput_2^b \mid \linput^{\neq b}], y^{\text{ILP}} \right)}_{\text{(b)}} \\
    &+ \lambda_2 \underbrace{\CE\left( P_\psi[\vinput_1^b \succ \vinput^{\neq b} \mid \linput^b], y^{\text{IVP}} \right)}_{\text{(c)}} \\
    &+ \lambda_2 \underbrace{\CE\left( P_\psi[\vinput_2^b \succ \vinput^{\neq b} \mid \linput^b], y^{\text{IVP}} \right)}_{\text{(c)}}
    \Big],
\end{aligned}
\label{eq:total_loss}
\end{equation}
where $\CE(\cdot, \cdot)$ is the cross-entropy loss, and $\lambda_1$ and $\lambda_2$ are balance weights of learning ILP and IVP.
By optimizing \eqref{eq:total_loss}, the \vl preference model outputs trajectory-level preference scores aligned with the \lc preference relations.

\section{Related Work}

\paragraph{Vision-Language Models for \RL.}

Our work is related to the literature on VLM rewards and preferences for embodied manipulation tasks~\citep{CLIP, R3M, LIV, VLM-RM, RL-VLM-F, CriticGPT}. These methods can be divided into three categories: (\romannumeral1) representation-based pre-training, (\romannumeral2) zero-shot inference, and (\romannumeral3) downstream fine-tuning.
For representation-based approaches, R3M~\citep{R3M} is pre-trained on the Ego4D dataset~\citep{Ego4D} to learn useful representations for downstream tasks. LIV~\citep{VIP}, which extends VIP~\citep{VIP} to multi-modal representations, is pre-trained on EpicKitchen dataset~\citep{EpicKitchen}, and can also be fine-tuned on target domain.
For zero-shot inference methods, VLM-RM~\citep{VLM-RM} utilizes CLIP~\citep{CLIP} as zero-shot vision-language rewards.
\RC~\citep{RoboCLIP} uses S3D~\citep{S3D}, which is pre-trained on HowTo100M dataset~\citep{HowTo100M}, as video-language model to compute vision-language reward with a single demonstration (a video or a text).
RL-VLM-F~\citep{RL-VLM-F} leverages Gemini-Pro~\citep{Gemini} and GPT-4V~\citep{GPT-4V} for zero-shot preference feedback.
CriticGPT~\citep{CriticGPT} is the representative method of (\romannumeral3), which fine-tunes multimodal LLMs on a instruction-following dataset, and utilizes the tuned model to provide preference feedback for downstream policy learning.
\ourmethod differs from these approaches that we do not suffer from burdensome training of (\romannumeral1) and (\romannumeral3), showing great computing efficiency. And \ourmethod learns more embodied manipulation knowledge compared with VLMs pre-trained on natural image-text data.

\begin{table*}[t]
\centering
\caption{Success rate of RLHF methods with scripted labels and \ourmethod labels. The results are reported with mean and standard deviation across five random seeds. The result of \ourmethod is \colorbox{\mycolor}{shaded} and is \textbf{bolded} if it exceeds or is comparable with that of RLHF approaches with scripted labels. \ourmethod Acc. denotes the accuracy of preference labels inferred by \ourmethod compared with scripted labels.}
\resizebox{0.85\linewidth}{!}{
\begin{tabular}{l|ll|ll|ll|l}
    \toprule
    {\bf Task} & {\bf P-IQL} & {\bf P-IQL+\ourmethod} & {\bf IPL} & {\bf IPL+\ourmethod} & {\bf CPL} & {\bf CPL+\ourmethod} & {\bf \ourmethod Acc.} \\
    \midrule
    Button Press
    & \mean{72.6} \std{7.1} 
    & \highlight \bmean{90.1} \std{3.9} 
    & \mean{50.6} \std{7.9} 
    & \highlight \bmean{56.0} \std{1.4} 
    & \mean{74.5} \std{8.2} 
    & \highlight \bmean{83.9} \std{11.8} 
    & \mean{93.0} 
    \\
    Door Close
    & \mean{79.2} \std{6.3} 
    & \highlight \bmean{79.2} \std{6.3} 
    & \mean{61.5} \std{9.4} 
    & \highlight \bmean{61.5} \std{9.4} 
    & \mean{98.5} \std{1.0} 
    & \highlight \bmean{98.5} \std{1.0} 
    & \mean{100.0} 
    \\
    Drawer Close
    & \mean{49.3} \std{4.2} 
    & \highlight \bmean{64.9} \std{2.9} 
    & \mean{64.3} \std{9.6} 
    & \highlight \mean{63.2} \std{4.7} 
    & \mean{45.6} \std{3.5} 
    & \highlight \bmean{57.5} \std{14.3} 
    & \mean{96.0} 
    \\
    Faucet Close
    & \mean{51.1} \std{7.5} 
    & \highlight \bmean{51.1} \std{7.5} 
    & \mean{45.4} \std{8.6} 
    & \highlight \bmean{45.4} \std{8.6} 
    & \mean{80.0} \std{2.9} 
    & \highlight \bmean{80.0} \std{2.9} 
    & \mean{100.0} 
    \\
    Window Open
    & \mean{62.4} \std{6.4} 
    & \highlight \bmean{69.7} \std{6.8} 
    & \mean{54.1} \std{6.7} 
    & \highlight \bmean{61.4} \std{8.6} 
    & \mean{91.6} \std{1.7} 
    & \highlight \bmean{99.1} \std{1.1} 
    & \mean{98.0} 
    \\
    \midrule
    {\bf Average}
    & \mean{62.9} 
    & \highlight \bmean{71.0} 
    & \mean{55.2} 
    & \highlight \bmean{57.5} 
    & \mean{78.0} 
    & \highlight \bmean{83.8} 
    & \mean{97.4} 
    \\
    \bottomrule
\end{tabular}
}
\label{tab:main_rlhf}
\end{table*}

\begin{table*}[!t]
\centering
\caption{Success rate of \ourmethod (i.e., P-IQL trained with \ourmethod labels) against IQL with VLM \textbf{rewards}. The results are reported with mean and standard deviation across five random seeds. The result of \ourmethod is \colorbox{\mycolor}{shaded} and the best score of all methods is \textbf{bolded}.}
\resizebox{0.9\linewidth}{!}{
\begin{tabular}{l|lllllll}
    \toprule
    {\bf Task} & {\bf R3M} & {\bf VIP} & {\bf LIV} & {\bf CLIP} & {\bf VLM-RM (0.0)} & {\bf VLM-RM (1.0)} & {\bf \ourmethod} \\
    \midrule
    Button Press
    & \mean{10.1} \std{2.3} 
    & \mean{68.4} \std{6.4} 
    & \mean{56.3} \std{1.9} 
    & \mean{59.5} \std{6.1} 
    & \mean{60.3} \std{6.1} 
    & \mean{64.3} \std{8.4} 
    & \highlight \bmean{90.1} \std{3.9} 
    \\
    Door Close
    & \mean{70.9} \std{5.3} 
    & \mean{74.8} \std{9.5} 
    & \mean{43.3} \std{3.2} 
    & \mean{43.6} \std{3.9} 
    & \mean{45.8} \std{8.5} 
    & \mean{41.1} \std{3.4} 
    & \highlight \bmean{79.2} \std{6.3} 
    \\
    Drawer Close
    & \mean{46.6} \std{2.6} 
    & \mean{70.4} \std{4.5} 
    & \mean{61.8} \std{5.7} 
    & \mean{69.4} \std{4.1} 
    & \mean{69.4} \std{4.5} 
    & \bmean{73.5} \std{5.4} 
    & \highlight \mean{64.9} \std{2.9} 
    \\
    Faucet Close
    & \mean{25.7} \std{23.6} 
    & \mean{40.9} \std{8.0} 
    & \mean{42.2} \std{6.3} 
    & \mean{59.6} \std{7.5} 
    & \bmean{60.1} \std{5.1} 
    & \mean{33.7} \std{15.3} 
    & \highlight \mean{51.1} \std{7.5} 
    \\
    Window Open
    & \mean{39.0} \std{6.6} 
    & \mean{42.7} \std{11.3} 
    & \mean{33.8} \std{6.4} 
    & \mean{26.4} \std{2.0} 
    & \mean{23.9} \std{1.9} 
    & \mean{23.7} \std{4.9} 
    & \highlight \bmean{69.7} \std{6.8} 
    \\
    \midrule
    {\bf Average}
    & \mean{38.5} 
    & \mean{59.4} 
    & \mean{47.5} 
    & \mean{51.7} 
    & \mean{51.9} 
    & \mean{47.3} 
    & \highlight \bmean{71.0} 
    \\
    \bottomrule
\end{tabular}
}
\label{tab:main_vlm}
\end{table*}

\paragraph{Preference-based Reinforcement Learning.}

\PbRL is a promising framework for aligning the agent with human values. However, feedback efficiency is a crucial challenge in \pbRL, with multiple recent studies striving to tackle. PEBBLE~\citep{PEBBLE} improves the efficiency by unsupervised pre-training. SURF~\citep{SURF} proposes to obtain pseudo labels using reward confidence. RUNE~\citep{RUNE} employs reward uncertainty to guide exploration. Meta-Reward-Net~\citep{Meta-Reward-Net} takes advantage of the performance of the Q-function as an additional signal to refine the accuracy of the reward model. \citet{hejna2023few} leverages meta-learning to pre-train the reward model, enabling fast adaptation to new tasks with few preference labels. SEER~\citep{SEER} enhances the efficiency of \pbRL by label smoothing and policy regularization. RAT~\citep{RAT} proposes to use \pbRL to attack deep RL agents.
Recently, a growing number of studies focus on offline \pbRL with the population of offline RL~\citep{levine2020offline, IQL, SEABO, OTDF}.
PT~\citep{PT} introduces a Transformer-based architecture for reward modeling. OPPO~\citep{OPPO} proposes to learn policies without a reward function. IPL~\citep{IPL} learns the Q-function from preferences, also eliminating the need of reward learning. CPL~\citep{CPL} further views \pbRL as a supervised learning problem, directly learning policies from preferences. FTB~\citep{FTB} introduces a diffusion model for better trajectory generation. PEARL~\citep{PEARL} proposes cross-task preference alignment to transfer preference labels between tasks and learn reward models robustly via reward distributional modeling. CAMP~\citep{CAMP} learns a diffusion-based preference model for preference alignment in multi-task RL~\citep{PiCor}.
\ourmethod addresses the labeling cost by learning a \vl preference model via \vl alignment, thereby providing generalized preferences to novel tasks.

\section{Experiments}
\label{sec:experiments}

In this section, we evaluate \ourmethod on \MW~\citep{Meta-World} benchmark and aim to answer the following questions:
\begin{itemize}[leftmargin=20pt]
    \item \textbf{Q1:} How do \ourmethod labels compare with scripted labels in offline RLHF? (Section~\ref{sec:exp_rlhf})
    \item \textbf{Q2:} How does \ourmethod compare with other vision-language rewards approaches? (Section~\ref{sec:exp_vlm})
    \item \textbf{Q3:} How does \ourmethod generalize to unseen tasks and language instructions? (Section~\ref{sec:exp_gen})
\end{itemize}

\begin{table*}[!t]
\centering
\caption{Success rate of \ourmethod (i.e., P-IQL trained with \ourmethod labels) against P-IQL with VLM \textbf{preferences} (denoted with prefix \textbf{P-}). The results are reported with mean and standard deviation across five random seeds. The result of \ourmethod is \colorbox{\mycolor}{shaded} and the best score of all methods is \textbf{bolded}.}
\resizebox{1\linewidth}{!}{
\begin{tabular}{l|llllllll}
    \toprule
    {\bf Task} & {\bf P-R3M} & {\bf P-VIP} & {\bf P-LIV} & {\bf P-CLIP} & {\bf P-VLM-RM (0.0)} & {\bf P-VLM-RM (1.0)}& {\bf \RC} & {\bf \ourmethod} \\
    \midrule
    Button Press
    & \mean{84.7} \std{5.8} 
    & \mean{41.2} \std{3.9} 
    & \mean{61.7} \std{5.1} 
    & \mean{62.9} \std{6.2} 
    & \mean{72.8} \std{5.0} 
    & \mean{44.2} \std{4.2} 
    & \mean{56.4} \std{7.3}
    & \highlight \bmean{90.1} \std{3.9} 
    \\
    Door Close
    & \mean{72.4} \std{11.5} 
    & \mean{54.2} \std{13.8} 
    & \mean{67.9} \std{6.3} 
    & \mean{53.3} \std{10.3} 
    & \mean{57.6} \std{2.9} 
    & \mean{45.7} \std{7.6} 
    & \mean{47.6} \std{6.7}
    & \highlight \bmean{79.2} \std{6.3} 
    \\
    Drawer Close
    & \mean{59.6} \std{6.5} 
    & \mean{63.0} \std{3.7} 
    & \mean{45.5} \std{10.4} 
    & \mean{63.4} \std{3.2} 
    & \mean{62.7} \std{3.0} 
    & \mean{49.2} \std{6.9} 
    & \bmean{73.0} \std{6.2}
    & \highlight \mean{64.9} \std{2.9} 
    \\
    Faucet Close
    & \mean{58.0} \std{4.5} 
    & \mean{51.1} \std{7.5} 
    & \bmean{62.3} \std{7.2} 
    & \mean{60.2} \std{10.4} 
    & \mean{57.3} \std{7.0} 
    & \mean{51.3} \std{9.5} 
    & \mean{62.1} \std{6.3}
    & \highlight \mean{51.1} \std{7.5} 
    \\
    Window Open
    & \mean{27.3} \std{5.0} 
    & \mean{50.2} \std{1.8} 
    & \mean{22.2} \std{18.1} 
    & \mean{28.4} \std{3.2} 
    & \mean{33.2} \std{5.4} 
    & \mean{20.7} \std{2.3} 
    & \mean{28.1} \std{4.6}
    & \highlight \bmean{69.7} \std{6.8} 
    \\
    \midrule
    {\bf Average}
    & \mean{60.4} 
    & \mean{51.9} 
    & \mean{51.9} 
    & \mean{53.6} 
    & \mean{56.7} 
    & \mean{42.2} 
    & \mean{53.4}
    & \highlight \bmean{71.0} 
    \\
    \bottomrule
\end{tabular}
}
\vspace{-0.5em}
\label{tab:main_vlm_p}
\end{table*}

\subsection{Setup}
\label{sec:exp_setup}

\paragraph{Implementation Details.}

We evaluate \ourmethod on the $5$ test tasks of \ourdata, including \emph{Button Press}, \emph{Door Close}, \emph{Drawer Close}, \emph{Faucet Close}, and \emph{Window Open}, while the other $45$ tasks of \MW~\citep{Meta-World} are used as training tasks. For implementing \ourmethod, we use the pre-trained ViT-B/16 CLIP model~\citep{CLIP} as our video encoder and language encoder. The weights of learning ILP and IVP in \eqref{eq:total_loss} are $\lambda_1 = 0.1$, $\lambda_2 = 0.5$, respectively. Additional hyperparameters of \ourmethod are detailed in Table~\ref{tab:VLP_params} in Appendix~\ref{app:exp_details}. All experiments are conducted on a single NVIDIA RTX 4090 GPU.

\subsection{How do \ourmethod labels compare with scripted labels in offline RLHF?}
\label{sec:exp_rlhf}

\paragraph{Baselines.}
We evaluate \ourmethod by combining it with recent offline RLHF algorithms:
(\romannumeral1) \textbf{P-IQL} (Preference IQL), which first learns a reward model from preferences and then learns a policy via IQL~\citep{IQL};
(\romannumeral2) \textbf{IPL}~\citep{IPL}, which learns a policy without reward learning by aligning the Q-function with preferences;
(\romannumeral3) \textbf{CPL}~\citep{CPL}, which directly learns a policy using a contrastive objective with maximum entropy principle, eliminating the need for reward learning and RL.

\paragraph{Evaluation.}
For each evaluation task, we train each RLHF method with scripted labels~\citep{christiano2017deep, PEBBLE} and \ourmethod labels (denoted as \textbf{+\ourmethod}), respectively. Scripted preference labels mean the preference labels computed based on the ground-truth rewards~\citep{christiano2017deep, PEBBLE}. The number of preference labels is set to $100$ for all tasks. The evaluation is conducted over $25$ episodes every $5000$ steps. Following~\citep{CPL}, we average the results of $8$ neighboring evaluations and take the maximum value among all averaged values as the result. Detailed hyperparameters of RLHF algorithms can be found in Appendix~\ref{app:exp_details}.

\paragraph{Results.}
Experimental results in Table~\ref{tab:main_rlhf} demonstrate that the performance of P-IQL+\ourmethod and CPL+\ourmethod is comparable with, and in some cases, outperforms that with scripted labels on all evaluation tasks. We hypothesize that the ground-truth reward of \emph{Button Press, Drawer Close} and \emph{Window Open} may not accurately represent the task goal~\citep{Text2Reward, Eureka, CARD}. However, by aligning video and language modalities through preference relations with language as conditions, the predicted \ourmethod labels directly represent how the video reflects the language instruction. Therefore, our method provides more accurate and preference labels and can generalize to unseen tasks.

\begin{table*}[!t]
\centering
\caption{The correlation coefficient of VLM rewards with ground-truth rewards and \ourmethod labels with scripted preference labels. Larger correlation means the predicted values are more correlated with the ground-truth.}
\resizebox{0.8\linewidth}{!}{
\begin{tabular}{l|rrrrrrr}
    \toprule
    {\bf Task} & {\bf R3M} & {\bf VIP} & {\bf LIV} & {\bf CLIP} & {\bf VLM-RM (0.0)} & {\bf VLM-RM (1.0)} & {\bf \ourmethod} \\
    \midrule
    Button Press
    & \mean{0.313} 
    & \mean{0.204} 
    & \mean{-0.281} 
    & \mean{0.127} 
    & \mean{0.153} 
    & \mean{-0.082} 
    & \highlight \bmean{0.581} 
    \\
    Door Close
    & \mean{0.735} 
    & \mean{0.125} 
    & \mean{0.600} 
    & \mean{-0.309} 
    & \mean{-0.152} 
    & \mean{-0.492} 
    & \highlight \bmean{1.000} 
    \\
    Drawer Close
    & \mean{-0.106} 
    & \mean{0.043} 
    & \mean{0.052} 
    & \mean{-0.151} 
    & \mean{-0.137} 
    & \mean{-0.031} 
    & \highlight \bmean{0.438} 
    \\
    Faucet Close
    & \mean{0.676} 
    & \mean{0.851} 
    & \mean{0.563} 
    & \mean{-0.301} 
    & \mean{-0.291} 
    & \mean{0.084} 
    & \highlight \bmean{1.000} 
    \\
    Window Open
    & \mean{0.411} 
    & \bmean{0.725} 
    & \mean{-0.568} 
    & \mean{0.336} 
    & \mean{0.405} 
    & \mean{-0.333} 
    & \highlight \mean{0.571} 
    \\
    \midrule
    {\bf Average}
    & \mean{0.406} 
    & \mean{0.390} 
    & \mean{0.073} 
    & \mean{-0.060} 
    & \mean{-0.005} 
    & \mean{-0.171} 
    & \highlight \bmean{0.718} 
    \\
    \bottomrule
\end{tabular}
}
\label{tab:main_corr}
\end{table*}

\begin{table*}[!ht]
\centering
\caption{The generalization abilities of our method on $5$ unseen tasks with different types of language instructions. Acc. denotes the accuracy of preference labels inferred by \ourmethod compared with ground-truth labels.}
\resizebox{0.73\linewidth}{!}{
\begin{tabular}{l|l|llll}
    \toprule
    {\bf Metric} & {\bf Seen} & {\bf Phrase} & {\bf Description} & {\bf Correct Color} & {\bf Incorrect Color} \\
    \midrule
    {\bf ITP Acc. ($\uparrow$)}
    & \mean{97.4} & \mean{95.8} & \mean{97.0} & \mean{97.0} & \mean{97.0} \\
    {\bf IVP Acc. ($\uparrow$)}
    & \mean{91.7} & \mean{90.5} & \mean{91.9} & \mean{91.9} & \mean{91.8} \\
    {\bf ILP Loss ($\downarrow$)}
    & \mean{0.705} & \mean{0.704} & \mean{0.704} & \mean{0.705} & \mean{0.705} \\
    {\bf Average Loss ($\downarrow$)}
    & \mean{0.555} & \mean{0.554} & \mean{0.558} & \mean{0.556} & \mean{0.557} \\
    \bottomrule
\end{tabular}
}
\label{tab:main_gen}
\end{table*}

\subsection{How does \ourmethod compare with other vision-language rewards approaches?}
\label{sec:exp_vlm}

\paragraph{Baselines.}
We compare \ourmethod with the following VLM rewards baselines:
(\romannumeral1) \textbf{R3M}~\citep{R3M}, which pre-trains visual representation by time-contrastive learning and \vl alignment;
(\romannumeral2) \textbf{VIP}~\citep{VIP}, which provides generalized visual reward and representation for downstream tasks via value-implicit pre-training;
(\romannumeral3) \textbf{LIV}~\citep{LIV}, which learns vision-language rewards and representation via multi-modal value pre-training;
(\romannumeral4) \textbf{CLIP}~\citep{CLIP}, which pre-trains by aligning vision-language representation on a large-scale image-text pairs dataset;
(\romannumeral5) \textbf{VLM-RM}~\citep{VLM-RM}, which provides zero-shot VLM rewards based on CLIP~\citep{CLIP}. VLM-RM includes a hyperparameter $\alpha$, which controls the goal-baseline regularization strength. In the evaluation, we denote the variant of $\alpha=0.0$ as \textbf{VLM-RM (0.0)} and the variant of $\alpha=1.0$ as \textbf{VLM-RM (1.0)}.
(\romannumeral6) \textbf{\RC}~\citep{RoboCLIP}, which provides zero-shot VLM rewards using pre-trained video-language models and a single demonstration (a video demonstration or a language description) of the task.

\paragraph{Evaluation.}
We first evaluate our method with the VLM baselines by directly training IQL with VLM \textbf{rewards}. \ourmethod is tested by training P-IQL with \ourmethod labels, and the experimental setting of our method is the same as that of Section~\ref{sec:exp_rlhf}. We further compare \ourmethod with VLM \textbf{preferences}, i.e., using predicted VLM rewards to compute preference labels for a fair comparison with our method. However, \RC obtains scalar trajectory-level rewards and we utilize them as trajectory return for preference labels calculation.
Implementation details of IQL and VLM baselines can be found in Appendix~\ref{app:exp_details}.

\paragraph{Results.}
Results in Table~\ref{tab:main_vlm} show that our method exceeds the VLM baselines that train IQL from VLM rewards by a large margin with an average success rate of \bmean{71.0}. As shown in Table~\ref{tab:main_vlm_p}, when the VLM baselines are trained with preferences computed by VLM rewards, our method still surpasses the baselines.
We further compute the preference label accuracy of each method, detailed in Table~\ref{tab:app_vlm_pref_acc}. The results show that \ourmethod exceeds VLM baselines, which do not learn relative relations of reward values.

\paragraph{Reward / Preference Correlation.}

To further investigate the advantages of \ourmethod model compared with VLM reward models, we compare the correlation between VLM rewards with ground-truth rewards and \ourmethod labels with scripted preference labels. Results in Table~\ref{tab:main_corr} indicate that \ourmethod labels exhibit a stronger correlation with scripted labels compared with VLM rewards.

\subsection{How does \ourmethod generalize to unseen tasks and language instructions?}
\label{sec:exp_gen}

\paragraph{Evaluation.}
We first evaluate how accurate $3$ kinds of \ourmethod labels are on the test tasks. We test the preference model with phrases, descriptions, and correct and incorrect object colors. Since the label of ILP is $0.5$ (i.e., two segments are equally preferred), we compute ILP loss with the (b) term in \eqref{eq:total_loss}, i.e., $- \sum_{b\in B} \CE\left( P_\psi[\vinput_1^b \succ \vinput_2^b \mid \linput^{\neq b}], y^{\text{ILP}} \right)$. Performance of ITP and IVP are measured with accuracy. Experimental details can be found in Appendix~\ref{app:exp_details}. 

\paragraph{Results.}
Table~\ref{tab:main_gen} shows that \ourmethod generalizes to unseen language instructions on unseen tasks with high ITP and IVP accuracy and low ILP loss.
However, using unseen phrases as language conditions leads to a performance drop, while unseen descriptions have a slight negative impact on ITP but a positive impact on IVP and ILP.
We think the reason is that phrases contain insufficient information about completing the task, while descriptions contain enough task information. \ourmethod generalizes well with suitable language information of tasks.
Also, \ourmethod exhibits strong generalization abilities on color.

\section{Conclusion}
\label{sec:conclusion}
In this paper, we propose \ourmethod, a novel \vl preference learning framework providing generalized preference feedback for embodied manipulation tasks. In our framework, we learn a \vl preference model via proposed \lc preference relations from the collected \vl preference dataset.
Experimental results on multiple simulated robotic manipulation tasks demonstrate that our method exceeds previous VLM rewards approaches and predicts accurate preferences compared with scripted labels. The results also show our method generalizes well to unseen tasks and unseen language instructions.

\section{Limitations}
\label{app:limitations}
In this paper, we focus on providing preferences for robotic manipulation tasks. First, \ourmethod is limited to the tasks that can be specified via videos and language instructions. While this covers a wide range of robotic tasks, certain tasks cannot be fully expressed via videos and language, such as complex assembly tasks requiring intricate spatial reasoning. Consequently, the risk of predicting incorrect preferences grows for complex tasks that are difficult to express. Second, if the language instruction lacks sufficient information of the task goal, the risk of giving incorrect labels still grows, as shown in Table~\ref{tab:main_gen}.

\bibliography{arxiv}

\clearpage
\appendix

\section{Experimental Details}
\label{app:exp_details}

\subsection{Tasks}

\paragraph{Meta-World.}

The tasks used in the experiments are from the test tasks of \ourdata. Figure~\ref{fig:tasks} shows these tasks and the task descriptions are as follows:

\begin{figure*}[!htbp]
\centering
\begin{tabular}{ccccc}
\subfloat[Button Press]{\includegraphics[width=0.17\linewidth]{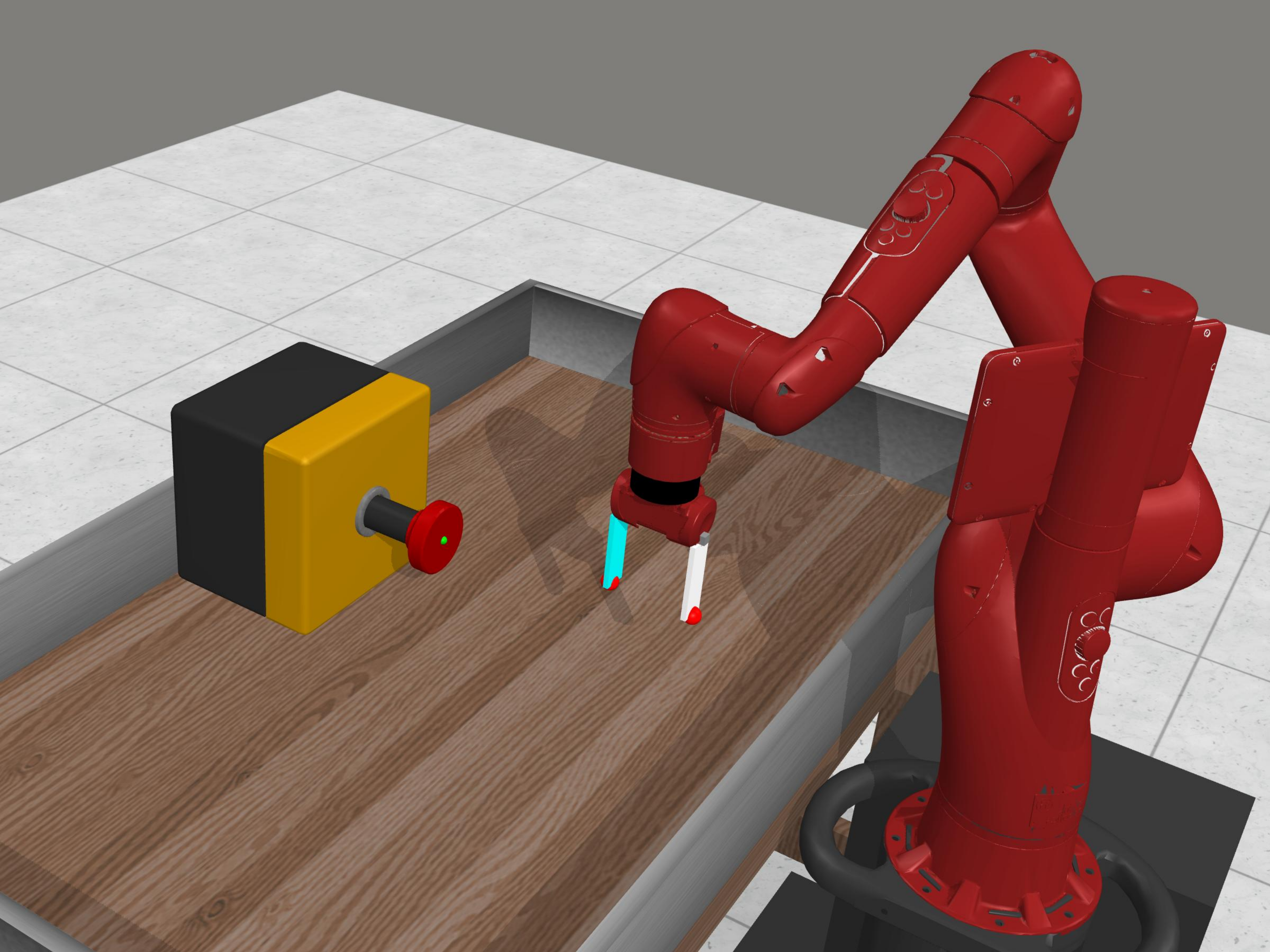}}
& \subfloat[Door Close]{\includegraphics[width=0.17\linewidth]{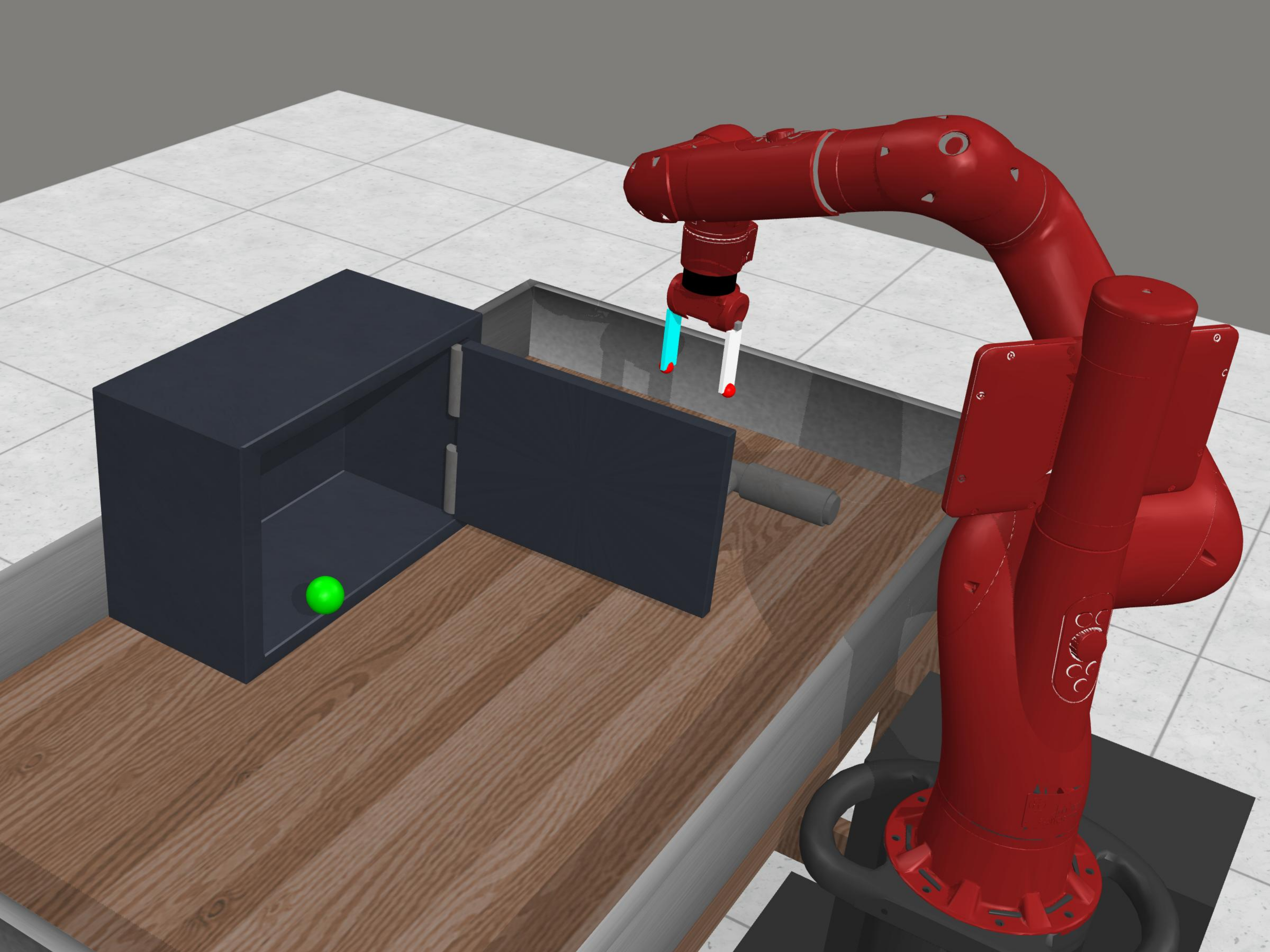}}
& \subfloat[Drawer Close]{\includegraphics[width=0.17\linewidth]{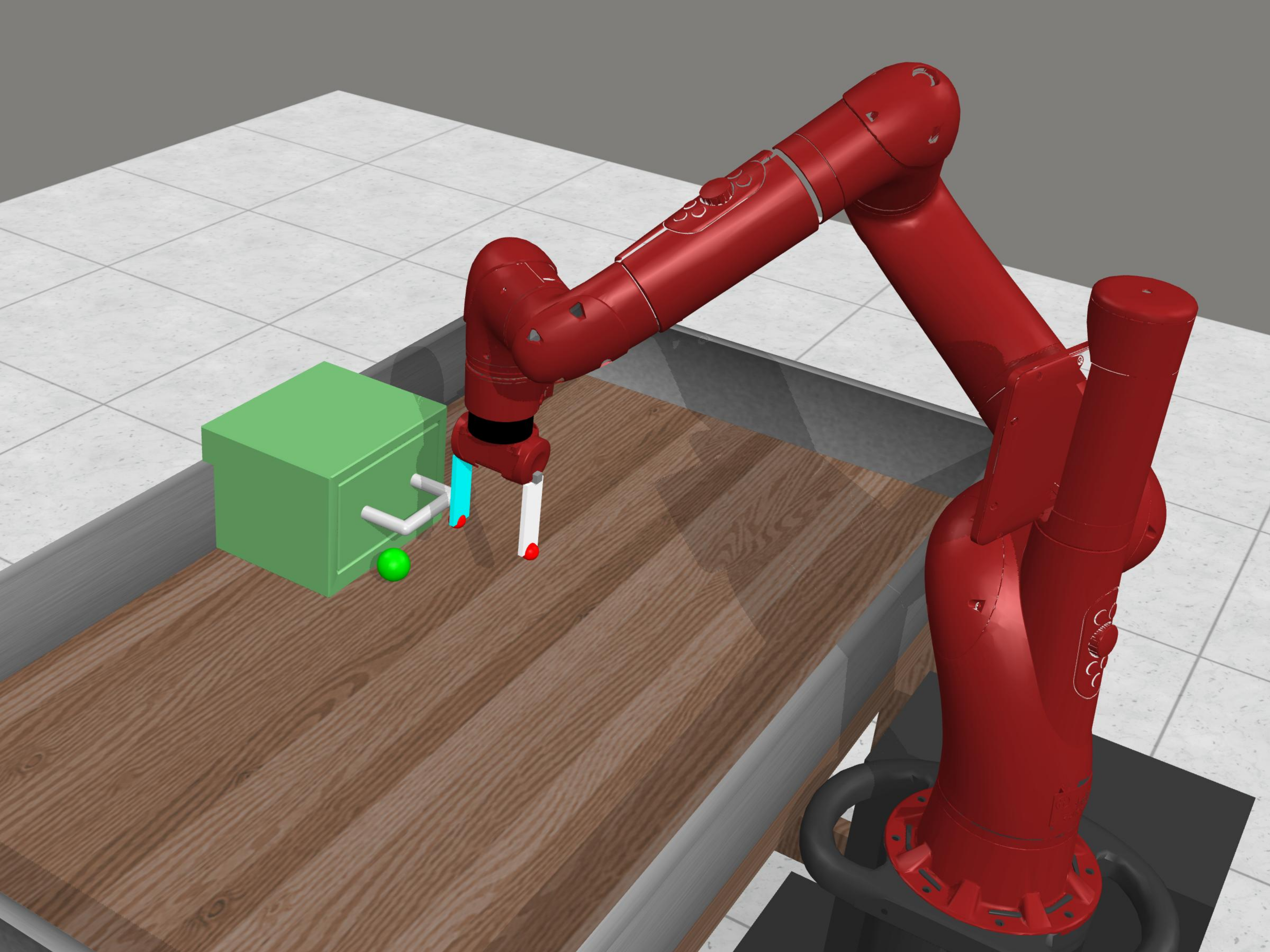}}
& \subfloat[Faucet Close]{\includegraphics[width=0.17\linewidth]{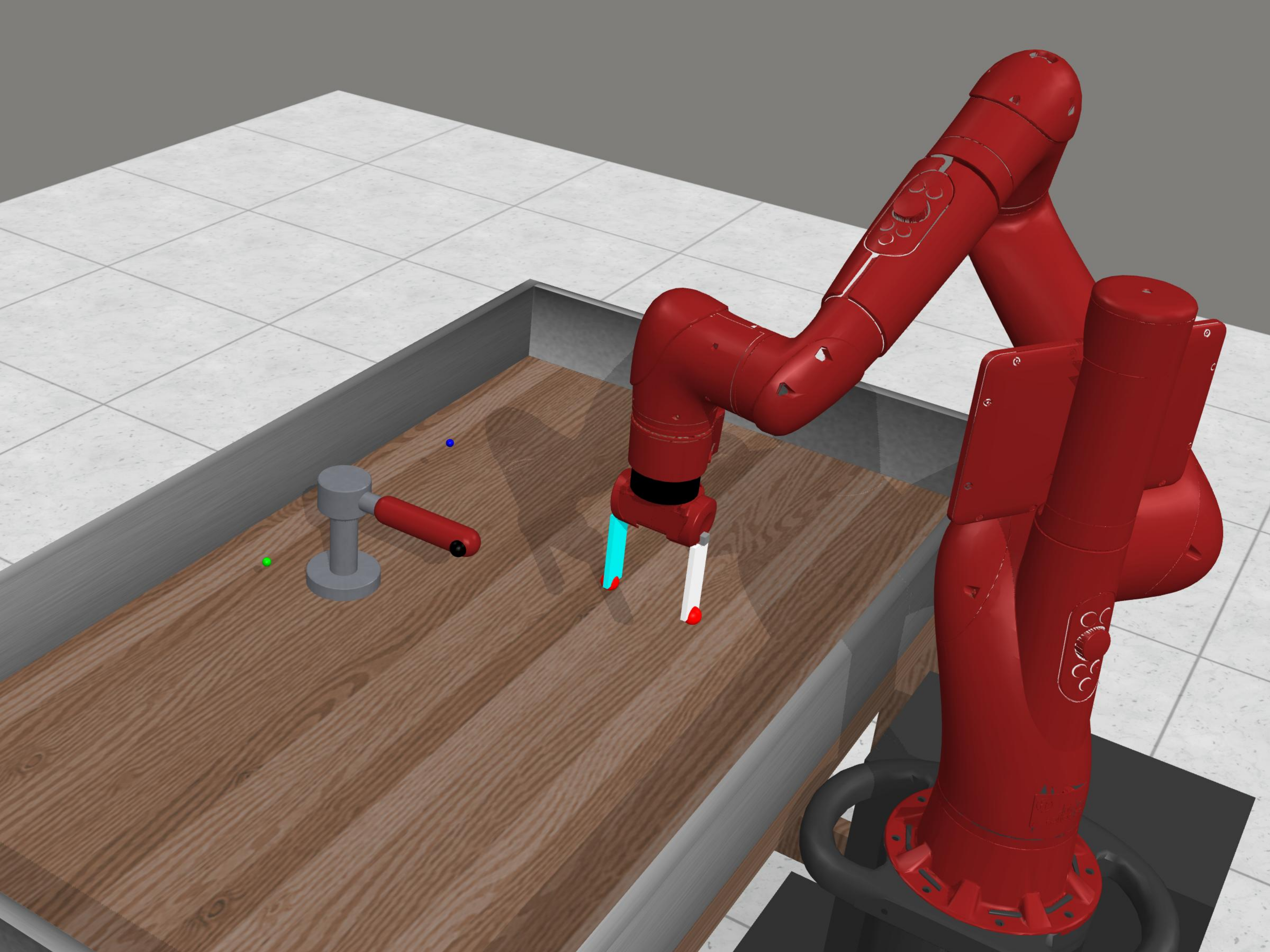}}
& \subfloat[Window Open]{\includegraphics[width=0.17\linewidth]{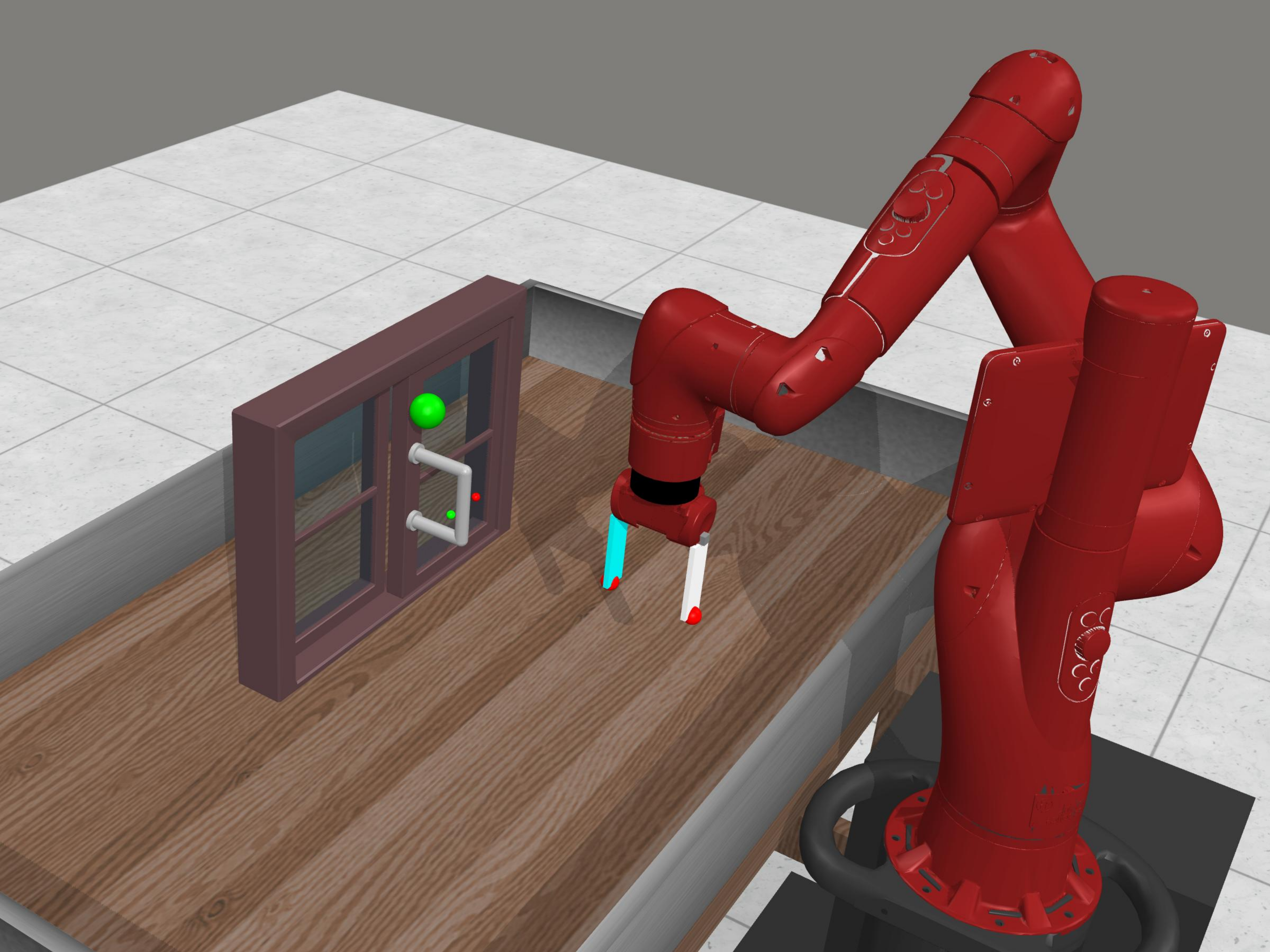}}
\end{tabular}
\caption{Five simulated robotic manipulation tasks used for experimental evaluation.}
\label{fig:tasks}
\end{figure*}

\begin{itemize} [leftmargin=20pt]
    \item Button Press: The goal of the robotic arm is to press the button. The initial position of the arm is randomly sampled.
    \item Door Close: The goal of the robotic arm is to close the door. The initial position of the arm is randomly sampled.
    \item Drawer Close: The goal of the robotic arm is to close the drawer. The initial position of the arm is randomly sampled.
    \item Faucet Close: The goal of the robotic arm is to close the faucet. The initial position of the arm is randomly sampled.
    \item Window Open: The goal of the robotic arm is to open the window. The initial position of the arm is randomly sampled.
\end{itemize}

\subsection{Implementation Details}

We implement our method based on the publicly released repository of LAPP~\citep{LAPP}.\footnote{\url{https://github.com/amberxie88/lapp}}
Following LAPP~\citep{LAPP}, we use a pre-trained ViT-B/16 CLIP~\citep{CLIP} model as our video encoder and language encoder. To achieve efficient learning, we uniformly sample $8$ frames to represent each video.
The detailed hyperparameters of our method are shown in Table~\ref{tab:VLP_params}. Training a \ourmethod model takes about $6$ hours on a single NVIDIA RTX 4090 GPU with $12$ CPU cores and $120$ GB memory, without costly pre-training process like VLM reward or VLM preference methods~\citep{R3M, VIP, LIV}.

\begin{table}[!htbp]
\centering
\caption{Hyperparameters of \ourmethod.}
\resizebox{0.9\linewidth}{!}{
\begin{tabular}{ll}
\toprule
\textbf{Hyperparameter} & \textbf{Value}  \\
\midrule
Prediction head & $(512, 256)$ \\
Number of self-attention layers & $2$ \\
Number of attention heads & $16$ \\
Batch size & $16$ \\
Optimizer & Adam \\
Learning rate & $3$e-$5$ \\
Learning rate decay & cosine decay \\
Weight decay & $0.1$ \\
Dropout & $0.1$ \\
Number of epochs & $15$k \\
Number of negative samples & $4$ \\
Number of video frames & $8$ \\
Weight of ILP loss $\lambda_1$ & $0.1$ \\
Weight of IVP loss $\lambda_1$ & $0.5$ \\
\bottomrule
\end{tabular}
}
\label{tab:VLP_params}
\end{table}

IQL, P-IQL, IPL and CPL are implemented based on the official repository of CPL and IPL.\footnote{\url{https://github.com/jhejna/cpl}}\footnote{\url{https://github.com/jhejna/inverse-preference-learning}} The hyperparameters of offline RL and RLHF algorithms are listed in Table~\ref{tab:common_params}, Table~\ref{tab:CPL_params}, and Table~\ref{tab:IPL_PIQL_params}. For the inference of \ourmethod labels, we first use K-means clustering to divide the trajectories of each test task into $2$ sets, following~\citet{PEARL}. Then we sample $100$ trajectory segments of length $50$ from each set to construct segment pairs and predict preference labels of these pairs with trained \ourmethod model. Training RL and RLHF algorithms take about $10$ minutes using a single NVIDIA RTX 4090 GPU with $6$ CPU cores and $60$ GB memory.

\begin{table}[!htbp]
\centering
\caption{Shared hyperparameters.}
\resizebox{1\linewidth}{!}{
\begin{tabular}{ll}
\toprule
\textbf{Hyperparameter} & \textbf{Value} \\
\midrule
Network architecture & $(256, 256)$ \\
Optimizer & Adam \\
Learning rate & $1$e-$4$ (CPL), $3$e-$4$ (IQL, IPL and P-IQL) \\
Batch size & $64$ \\
Discount & $0.99$ \\
Dropout & $0.25$ \\
Training steps & $100000$ \\
\midrule
Segment length & $50$ (RLHF) \\
Number of queries & $100$ (RLHF) \\
Temperature & $0.3333$ (IQL, IPL and P-IQL) \\
Expectile & $0.7$ (IQL, IPL and P-IQL) \\
Soft target update rate & $0.005$ (IQL, IPL and P-IQL) \\
\bottomrule
\end{tabular}
}
\label{tab:common_params}
\end{table}

\begin{table}[!htbp]
\centering
\caption{Hyperparameters of CPL.}
\resizebox{0.55\linewidth}{!}{
\begin{tabular}{ll}
\toprule
\textbf{Hyperparameter} & \textbf{Value} \\
\midrule
Temperature & $0.1$ \\
Contrastive bias & $0.5$ \\
BC weight & $0.0$ \\
BC steps & $10000$ \\
\bottomrule
\end{tabular}
}
\label{tab:CPL_params}
\end{table}

\begin{table}[!htbp]
\centering
\caption{Hyperparameters of IPL and P-IQL.}
\resizebox{0.75\linewidth}{!}{
\begin{tabular}{ll}
\toprule
\textbf{Hyperparameter} & \textbf{Value} \\
\midrule
Regularization weight (IPL) & $0.5$ \\
Reward learning steps (P-IQL) & $30$ \\
\bottomrule
\end{tabular}
}
\label{tab:IPL_PIQL_params}
\end{table}

For VLM methods, R3M, VIP, LIV, VLM-RM, and RoboCLIP are implemented based on their official repositories.\footnote{\url{https://github.com/facebookresearch/r3m}}\footnote{\url{https://github.com/facebookresearch/vip}}\footnote{\url{https://github.com/penn-pal-lab/LIV}}\footnote{\url{https://github.com/AlignmentResearch/vlmrm}}\footnote{\url{https://github.com/sumedh7/RoboCLIP}}
The CLIP baseline is a variant of VLM-RM and is implemented based on the code of VLM-RM. The language inputs of the VLM baselines except are as listed in Table~\ref{tab:VLM_lang}. R3M, LIV, CLIP, and \RC only require the target column as language inputs, while VLM-RM additionally needs a baseline as a regularization term. R3M requires an initial image and we use the first frame of each trajectory as the initial image, while VIP requires a goal image for VLM rewards inference and we use the last frame of expert videos.

\begin{table}[!htbp]
\centering
\caption{Language inputs used for evaluating VLM baselines on the test tasks.}
\resizebox{1\linewidth}{!}{
\begin{tabular}{l|ll}
    \toprule
    {\bf Task} & {\bf Target} & {\bf Baseline (for VLM-RM)} \\
    \midrule
    Button Press & press button & button \\
    Door Close & close door & door \\
    Drawer Close & close drawer & drawer \\
    Faucet Close & turn faucet left & faucet \\
    Window Open & move window left & window \\
    \bottomrule
\end{tabular}
}
\label{tab:VLM_lang}
\end{table}

\section{Additional Experimental Results}
\label{app:exp_results}

\paragraph{Evaluation on \MS Tasks.}

To examine the effects of \ourmethod on more challenging tasks, we conduct experiments on \MS~\citep{ManiSkill2} benchmark. We leverage \emph{MoveBucket-v1}, \emph{OpenCabinetDrawer-v1}, \emph{PegInsertionSide-v0}, \emph{PickCube-v0}, \emph{PickSingleEGAD-v0}, \emph{PlugCharger-v0}, \emph{StackCube-v0}, and \emph{TurnFaucet-v0} as training tasks and evaluate \ourmethod on \emph{LiftCube-v0}, \emph{OpenCabinetDoor-v1}, \emph{PushChair-v1} tasks. Table~\ref{tab:maniskill} summarizes the average \ourmethod label accuracy on the three test tasks compared to scripted labels and the results demonstrate the strong generalization capabilities of \ourmethod.

\begin{table}[!htbp]
\centering
\caption{Preference label accuracy of \ourmethod on \MS test tasks.}
\resizebox{0.65\linewidth}{!}{
\begin{tabular}{l|l}
    \toprule
    {\bf Task} & {\bf \ourmethod Acc.} \\
    \midrule
    LiftCube-v0 & \mean{100.0} \\
    OpenCabinetDoor-v1 & \mean{100.0} \\
    PushChair-v1 & \mean{93.8} \\
    \midrule
    {\bf Average} & \mean{97.9} \\
    \bottomrule
\end{tabular}
}
\label{tab:maniskill}
\end{table}

\paragraph{Attention Map Visualization.}

\begin{figure}[!htbp]
\vspace{-1em}
\centering
\begin{tabular}{c}
\hspace{-0.2em} \subfloat[Drawer Close (Shift closer and secure the drawer shut)]{\includegraphics[width=0.7\linewidth]{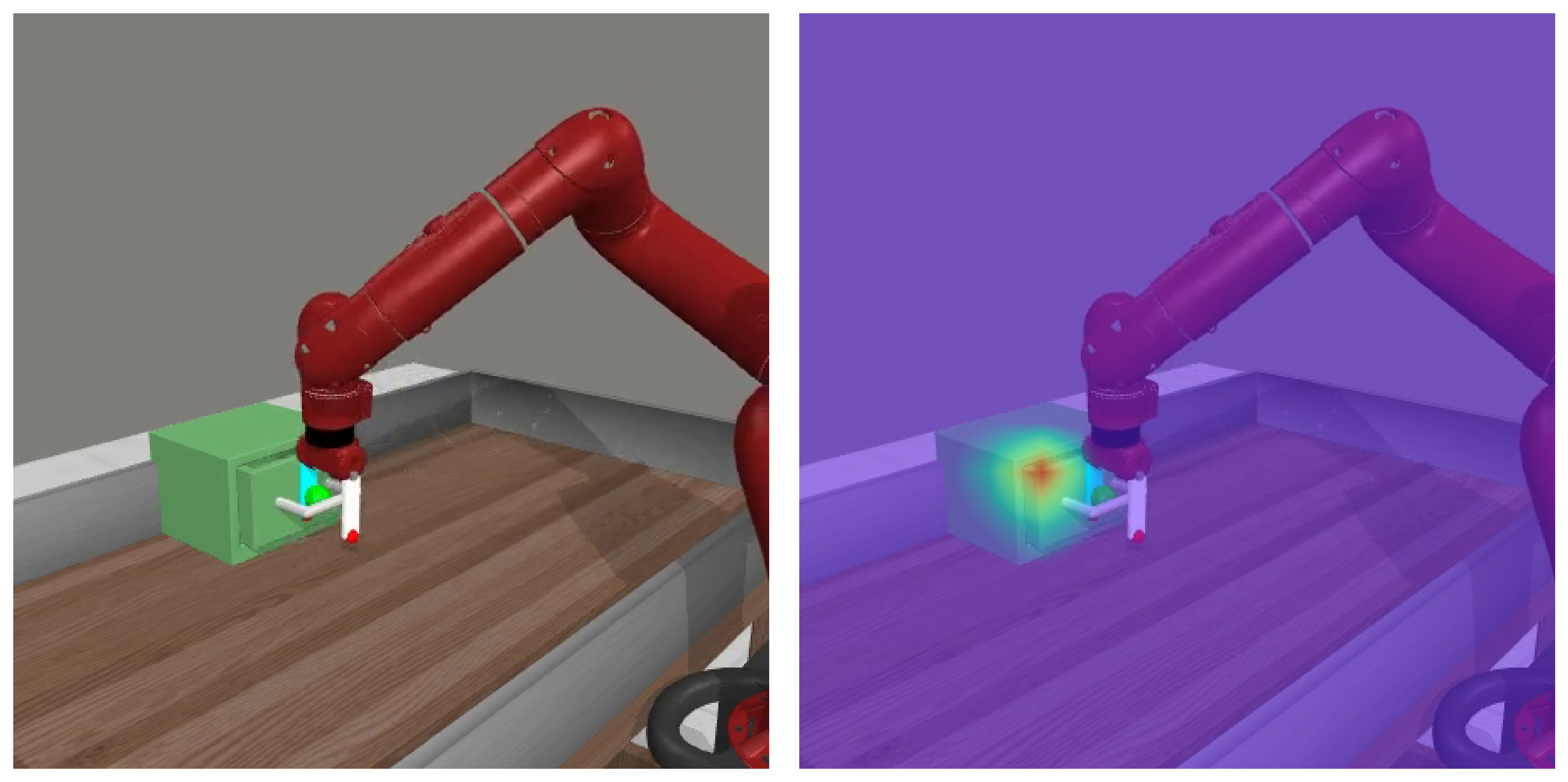}} \\
\hspace{-0.2em} \subfloat[Door Close (Direct the gripper to the door handle and press to seal it)]{\includegraphics[width=0.7\linewidth]{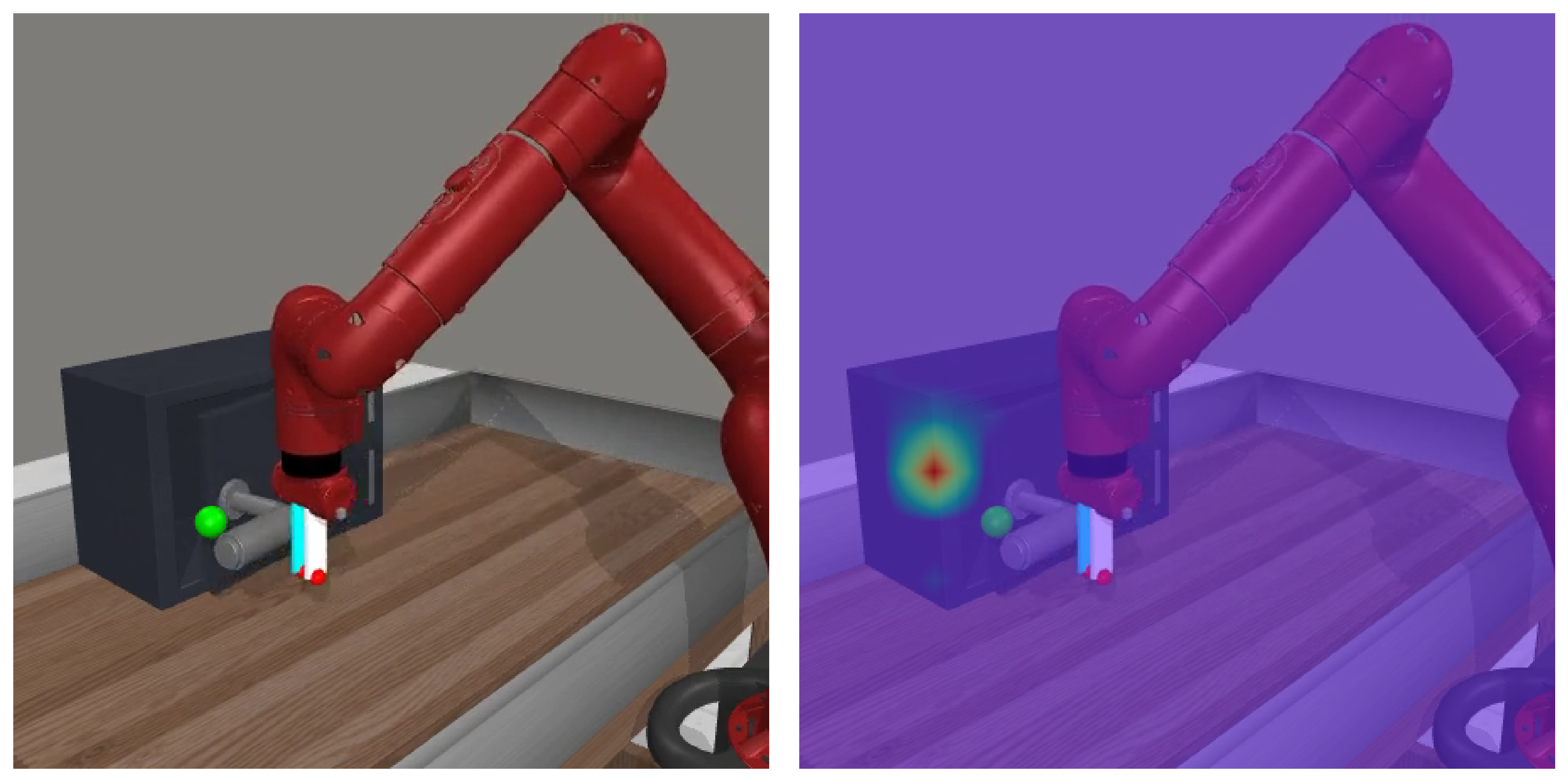}}
\end{tabular}
\caption{Attention map visualization of \emph{Drawer Close} and \emph{Door Close}. The language instruction is shown at the bottom of each subfigure.}
\vspace{-1em}
\label{fig:attn_map}
\end{figure}

We further analyze \ourmethod by visualizing the attention maps of the cross-attention. Results in Figure~\ref{fig:attn_map} show that regions of the objects related to language instructions exhibit high attention weights. For example, in the Drawer Close task, our \vl preference model specifically focuses on whether the drawer is closed, with the attention map highlighting the edges of the drawer to monitor its position and similarly for Door Close task. These observations demonstrate that our \vl preference model effectively learns to guide language tokens to attend to relevant regions in the videos and illustrate the effectiveness of our cross-attention mechanism in bridging vision and language modalities for precise task understanding.

\paragraph{Effects of $\lambda_1$ and $\lambda_2$.}
$\lambda_1$ and $\lambda_2$ in \eqref{eq:total_loss} control the strength of ILP and IVP learning, respectively. To investigate how $\lambda_1$ and $\lambda_2$ influence \ourmethod, we conduct experiments by vary $\lambda_1$ across $\{0.0, 0.1, 0.5\}$ and $\lambda_2$ across $\{0.0, 0.5, 1.0\}$. Results in Table~\ref{tab:abla_loss} show that the performance of \ourmethod drops with too small or too large $\lambda_1$. Meanwhile, without IVP learning (i.e., $\lambda_2=0$), the performance of IVP and ILP significantly decreases. We speculate that IVP is crucial for \lc preference learning. Without IVP learning, the learned \ourmethod model degenerates into a vanilla preference model without language as conditions.

\begin{table}[!htbp]
\centering
\caption{Accuracy of \ourmethod labels with different loss. Acc. denotes the accuracy of preference labels inferred by \ourmethod compared with ground-truth labels.}
\resizebox{1\linewidth}{!}{
\begin{tabular}{ll|llll}
    \toprule
    {\bf $\lambda_1$} & {\bf $\lambda_2$} & {\bf ITP Acc. ($\uparrow$)} & {\bf IVP Acc. ($\uparrow$)} & {\bf ILP Loss ($\downarrow$)} & {\bf Avg. Loss ($\downarrow$)} \\
    \midrule
    \mean{0.0} & \mean{0.5} & \mean{95.4} & \mean{74.1}
    & \mean{0.728} & \mean{0.618} \\
    \mean{0.5} & \mean{0.5} & \mean{85.8} & \mean{74.7}
    & \mean{0.702} & \mean{0.578} \\
    \midrule
    \mean{0.1} & \mean{0.0} & \mean{96.2} & \mean{63.0}
    & \mean{0.775} & \mean{0.646} \\
    \mean{0.1} & \mean{1.0} & \mean{95.8} & \mean{96.5}
    & \mean{0.699} & \mean{0.554} \\
    \midrule
    \rowcolor{\mycolor}
    \mean{0.1} & \mean{0.5} & \mean{97.4} & \mean{91.7}
    & \mean{0.705} & \mean{0.555} \\
    \bottomrule
\end{tabular}
}
\label{tab:abla_loss}
\end{table}

\paragraph{Effects of Preference Dataset Size.}
We investigate how the preference dataset size influences our method. We conduct additional experiments by varying the dataset size across $\{50\%, 75\%, 100\%\}$. Results in Table~\ref{tab:abla_size} indicate that the performance of VLP downgrades as the dataset size decreases.

\begin{table}[!htbp]
\centering
\caption{Accuracy of \ourmethod labels with different data size. Acc. denotes the accuracy of preference labels inferred by \ourmethod compared with ground-truth labels.}
\resizebox{1\linewidth}{!}{
\begin{tabular}{l|llll}
    \toprule
    {\bf Data} & {\bf ITP Acc. ($\uparrow$)} & {\bf IVP Acc. ($\uparrow$)} & {\bf ILP Loss ($\downarrow$)} & {\bf Avg. Loss ($\downarrow$)} \\
    \midrule
    \mean{50\%} & \mean{94.2} & \mean{89.6}
    & \mean{0.699} & \mean{0.557} \\
    \mean{75\%} & \mean{95.2} & \mean{89.7}
    & \mean{0.707} & \mean{0.555} \\
    \rowcolor{\mycolor}
    \mean{100\%} & \mean{97.4} & \mean{91.7}
    & \mean{0.705} & \mean{0.555} \\
    \bottomrule
\end{tabular}
}
\label{tab:abla_size}
\end{table}

\paragraph{Preference Label Accuracy.}
To compare the relative relation of VLM rewards with \ourmethod, we compute the preference label accuracy of each method. The accuracy is measured by comparing the predicted preference labels with scripted preference labels.
The results in Table~\ref{tab:app_vlm_pref_acc} show that \ourmethod exceeds the VLM baselines by a large margin, demonstrating VLM rewards do not capture the relative reward relationship.

\begin{table*}[!htbp]
\centering
\caption{Preference label accuracy of \ourmethod against VLM baselines. The accuracy of our method is \colorbox{\mycolor}{shaded} and the best score of all methods is \textbf{bolded}.}
\resizebox{0.9\linewidth}{!}{
\begin{tabular}{l|llllllll}
    \toprule
    {\bf Task} & {\bf R3M} & {\bf VIP} & {\bf LIV} & {\bf CLIP} & {\bf VLM-RM (0.0)} & {\bf VLM-RM (1.0)} & {\bf \RC} & {\bf \ourmethod} \\
    \midrule
    Button Press
    & \mean{91.0} 
    & \mean{40.0} 
    & \mean{62.0} 
    & \mean{53.0} 
    & \mean{62.0} 
    & \mean{41.0} 
    & \mean{46.0}
    & \highlight \bmean{93.0} 
    \\
    Door Close
    & \mean{98.0} 
    & \mean{57.0} 
    & \mean{97.0} 
    & \mean{49.0} 
    & \mean{59.0} 
    & \mean{10.0} 
    & \mean{61.0}
    & \highlight \bmean{100.0} 
    \\
    Drawer Close
    & \mean{66.0} 
    & \mean{49.0} 
    & \mean{39.0} 
    & \mean{66.0} 
    & \mean{65.0} 
    & \mean{58.0} 
    & \mean{43.0}
    & \highlight \bmean{96.0} 
    \\
    Faucet Close
    & \mean{98.0} 
    & \bmean{100.0} 
    & \mean{97.0} 
    & \mean{38.0} 
    & \mean{25.0} 
    & \mean{65.0} 
    & \mean{63.0}
    & \highlight \bmean{100.0} 
    \\
    Window Open
    & \mean{72.0} 
    & \mean{88.0} 
    & \mean{16.0} 
    & \mean{81.0} 
    & \mean{88.0} 
    & \mean{16.0} 
    & \mean{49.0}
    & \highlight \bmean{98.0} 
    \\
    \midrule
    {\bf Average}
    & \mean{85.0} 
    & \mean{66.8} 
    & \mean{62.2} 
    & \mean{57.4} 
    & \mean{59.8} 
    & \mean{38.0} 
    & \mean{52.4}
    & \highlight \bmean{97.4} 
    \\
    \bottomrule
\end{tabular}
}
\label{tab:app_vlm_pref_acc}
\end{table*}

\paragraph{Different VLMs/LLMs for Language Instruction Generation.}
To see the influence of different language model on our method, we we conduct additional experiments using instructions from less capable model, such as GPT-3.5 and open-source Llama-3.1-8B-Instruct.
We observe that generating diverse language instructions does not necessarily require strong VLMs like GPT-4V, even open-source Llama-3.1-8B-Instruct can accomplish this job since the language model is prompted with a diverse set of examples, following LAMP~\citep{LAMP}. The results in Table~\ref{tab:app_llm} show that the model's performance is relatively stable across different LLMs.

\begin{table}[!htbp]
\centering
\caption{Preference label accuracy of \ourmethod with language instructions generated by different VLMs/LLMs.}
\resizebox{1\linewidth}{!}{
\begin{tabular}{l|lll}
    \toprule
    {\bf Task} & {\bf GPT-4V} & {\bf GPT-3.5} & {\bf Llama-3.1-8B-Inst.} \\
    \midrule
    Button Press
    & \mean{93.0} 
    & \mean{93.0}
    & \mean{91.0}
    \\
    Door Close
    & \mean{100.0} 
    & \mean{100.0}
    & \mean{98.0}
    \\
    Drawer Close
    & \mean{96.0} 
    & \mean{96.0}
    & \mean{97.0}
    \\
    Faucet Close
    & \mean{100.0} 
    & \mean{100.0}
    & \mean{100.0}
    \\
    Window Open
    & \mean{98.0} 
    & \mean{99.0}
    & \mean{99.0}
    \\
    \midrule
    {\bf Average}
    & \mean{97.4} 
    & \mean{97.6}
    & \mean{97.0}
    \\
    \bottomrule
\end{tabular}
}
\label{tab:app_llm}
\end{table}

\section{Details of \ourdata Collection}
\label{app:data_details}

For the $50$ robotic manipulation tasks in \MW~\citep{Meta-World}, we divide \emph{Button Press}, \emph{Door Close}, \emph{Drawer Close}, \emph{Faucet Close}, and \emph{Window Open} as test tasks and the other $45$ tasks as train tasks.
For each task, we leverage scripted policies of \MW~\citep{Meta-World} to collect trajectories. For expert trajectories, we add Gaussian noise sampled from $\gN (0, 0.1)$. For medium trajectories, we utilize the \texttt{near\_object} flag returned by each task to determine whether the first subtask is completed and add Gaussian noise sampled from $\gN (0, 0.5)$. For random trajectories, the actions are sampled from uniform distribution $\gU [0, 1]$.
We collect $32$ trajectories of each type of trajectory for each task, resulting in a total of $4800$ trajectories for all tasks.
We query GPT-4V~\citep{GPT-4V} to generate language instructions by the prompt containing an example of generating diverse language instructions, an example of generating synonym nouns, task name, task instruction, and an image rendering the task. The detailed prompt we used is shown in Table~\ref{tab:detailed_gpt_prompt}.

\begin{table*}[!htbp]
\centering
\caption{
Prompt for generating diverse language instructions. 
The verb structures list and synonym nouns example are from Table 2 and Table 4 in LAMP~\citep{LAMP}, respectively.
}
\begin{tcolorbox}
    \raggedright
    \small
\texttt{System Message:} Suppose you are an advanced visual assistant. Your task is to generate more instructions with the same meaning but different expressions based on the task instruction I provide, generating 40 new instructions for each task. The instructions you generate need to be as simple and clear as possible. Below is an example of an answer for picking up an object. The answer should be formatted as a Python list.

-- Begin of instruction example -- \\
\texttt{Task instruction}: "Pick up the [NOUN]" \\
\texttt{Answer}: \\
\texttt{Verb Structures List} \\
-- End of instruction example -- \\

Moreover, you need to be mindful to replace the nouns in the instructions with synonyms, such as replacing "bag" with the following words in the Python list: \\
-- Begin of synonym example -- \\
\texttt{Synonym Nouns} \\
-- End of synonym example -- \\

The tasks are from Meta-World benchmark and the image of the task is rendered in a 3D simulation environment. In the environment, there is a wooden table and a robotic arm. The robotic arm is placed above the table. The robotic arm needs to manipulate the object(s) on the table to complete tasks. \\
My instruction for \texttt{Task Name} task: \texttt{Task Instruction} \\
\texttt{Answer}: \\
\end{tcolorbox}

\label{tab:detailed_gpt_prompt}
\end{table*}

\section{Discussions}

\paragraph{How do ILP and IVP benefit \ourmethod?}
The inclusion of ILP and IVP in our training data serves critical roles in enhancing the generalization and robustness of our model. ILP allows our model to learn to disregard language variations when they do not impact the preference outcomes, thus training the model to focus on task-relevant features rather than linguistic discrepancies. On the other hand, IVP facilitates the model's ability to generalize across different tasks by learning to associate videos with their corresponding task-specific language instructions effectively. This capability is crucial when the model encounters new tasks or language contexts, as it must discern relevant from irrelevant information to make accurate preference predictions. By training with both ILP and IVP, our model learns a more holistic understanding of the task space, which not only improves its performance on seen tasks but also enhances its adaptability to new, unseen tasks or variations in task descriptions, as evidenced by our experimental results where the model demonstrated generalization capabilities.

\paragraph{How does different train-test split influence \ourmethod?}
We conduct experiments on the \MW ML45 benchmark, training the \vl preference model on its training tasks and evaluating on its test tasks. We compute \ourmethod label accuracy by comparing \ourmethod label with scripted preference labels. The results shown in Table~\ref{tab:app_ml45} demonstrate the strong generalization capability of our method on unseen tasks in ML45. This reinforces the robustness and adaptability of our framework regardless of task split.

\begin{table}[!htbp]
\centering
\caption{Preference label accuracy of \ourmethod on ML45 test tasks.}
\resizebox{0.5\linewidth}{!}{
\begin{tabular}{l|l}
    \toprule
    {\bf Task} & {\bf \ourmethod Acc.} \\
    \midrule
    Bin Picking & \mean{95.0} \\
    Box Close & \mean{90.0} \\
    Door Lock & \mean{100.0} \\
    Door Unlock & \mean{100.0} \\
    Hand Insert & \mean{100.0} \\
    \midrule
    {\bf Average} & \mean{97.0} \\
    \bottomrule
\end{tabular}
}
\label{tab:app_ml45}
\end{table}

\end{document}